\documentclass{article}

\usepackage{microtype}
\usepackage{graphicx}
\usepackage{subcaption}
\usepackage{booktabs} 

\usepackage{hyperref}

\usepackage[preprint]{icml2026}

\usepackage{amsmath}
\usepackage{amssymb}
\usepackage{mathtools}
\usepackage{amsthm}
\usepackage{xspace}
\usepackage{enumitem}
\usepackage{listings}
\usepackage{afterpage}
\usepackage{float}

\usepackage{fontawesome5}

\definecolor{chipthinkfill}{HTML}{FBE1D5}
\definecolor{chipthinkline}{HTML}{EB6834}
\definecolor{chiptailfill}{HTML}{D6E5F8}
\definecolor{chiptailline}{HTML}{2A78D6}
\definecolor{grouplabel}{HTML}{8A8984}
\newcommand{\thinkchip}{{\setlength{\fboxsep}{0.7pt}\setlength{\fboxrule}{0.5pt}\fcolorbox{chipthinkline}{chipthinkfill}{\makebox[0.8em]{\color{chipthinkline}\scriptsize\faBrain}}}}
\newcommand{\tailchip}{{\setlength{\fboxsep}{0.7pt}\setlength{\fboxrule}{0.5pt}\fcolorbox{chiptailline}{chiptailfill}{\makebox[0.8em]{\color{chiptailline}\scriptsize\faTerminal}}}}
\newcommand{\nochip}{\phantom{\thinkchip}}
\newcommand{\grouplabel}[1]{\textcolor{grouplabel}{\footnotesize\itshape #1}}

\usepackage[capitalize,noabbrev]{cleveref}

\theoremstyle{plain}

\theoremstyle{definition}

\theoremstyle{remark}

\newcommand{\method}{Muted Distillation\xspace}
\newcommand{\methodshort}{MD\xspace}
\newcommand{\eot}{\texttt{</think>}\xspace}

\newif\ifshowcomments
\showcommentstrue
\ifshowcomments
  \newcommand{\cmt}[1]{{\color{red}[#1]}}
  
\else
  \newcommand{\cmt}[1]{\unskip}
  
\fi

\icmltitlerunning{Muted Distillation: Sealing Reasoning from Teacher Unlocks Token-Efficient Agents}

\begin{document}

\twocolumn[{
  \icmltitle{Muted Distillation: Sealing Reasoning from Teacher \texorpdfstring{\\}{}
    Unlocks Token-Efficient Agents}


  \icmlsetsymbol{equal}{*}

  \begin{icmlauthorlist}
    \icmlauthor{Yuwei Zhang}{ext1}
    \icmlauthor{Ian Miao}{aff}
    \icmlauthor{Wentao Zhang}{ext2}
    \icmlauthor{Chen Qu}{aff}
    \icmlauthor{Bill Bi}{aff}
    \icmlauthor{Jingbo Shang}{ext1}
    \icmlauthor{Donna Schut}{aff}
  \end{icmlauthorlist}

  \icmlaffiliation{aff}{Google}
  \icmlaffiliation{ext1}{UC San Diego}
  \icmlaffiliation{ext2}{University of Waterloo}

  \icmlcorrespondingauthor{Yuwei Zhang}{yuz163@ucsd.com}
  \icmlcorrespondingauthor{Ian Miao}{yilinmiao@google.com}
  \icmlcorrespondingauthor{Chen Qu}{cqu@google.com}

  \icmlkeywords{On-policy Distillation, Knowledge-Work Agents, Token Efficiency, Large Language Models}

  \vskip 0.15in
  \begin{center}
    \includegraphics[width=0.88\textwidth]{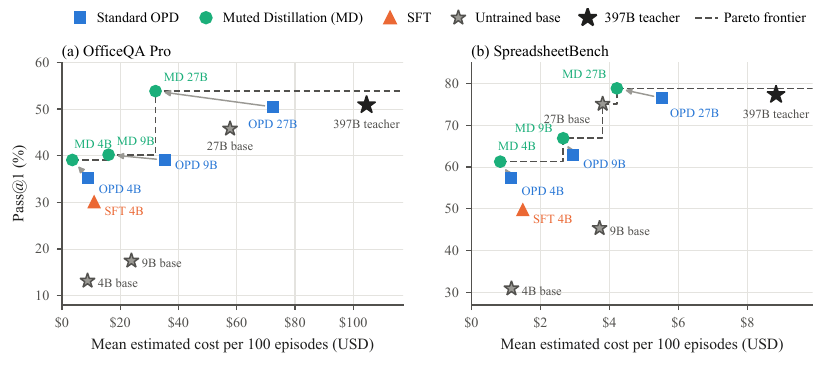}
    \captionof{figure}{Pass@1 against the mean estimated inference cost per 100 episodes for Qwen3.5 students of 4B, 9B, and 27B parameters, each trained recipe at its best evaluated checkpoint.
    Cost counts uncached prompt and output tokens at the median verified paid-provider rate for each backbone (rates frozen on 2026-09-10).}
    \label{fig:pareto}
  \end{center}
  \vskip 0.2in
}]

\printAffiliationsAndNotice{Work done while the first and third authors are at Google.}

\begin{abstract}
On-policy distillation (OPD) is commonly adopted as one of the components in post-training of large language models (LLMs), but in multi-turn agentic tasks we find that the student model increases their reasoning lengths after training.
Our empirical analysis shows that the reasoning token lengths is affected by how much the student can distract the teacher's scores.
We introduce \method (\methodshort), a simple variant of OPD that hides the student's reasoning from the teacher, so the teacher judges what the student does without the interference of student's reasonings. \methodshort takes a few lines of code and adds no extra data or hyperparameters. On the knowledge-work benchmarks OfficeQA Pro and SpreadsheetBench, it reduces output tokens per episode by 57\% and 53\% compared with standard OPD, making the student more concise than it was before training. At the same time, it matches OPD's accuracy on OfficeQA Pro and improves pass@1 by 4.8 points on SpreadsheetBench.
\end{abstract}

\section{Introduction}

On-policy Distillation (OPD) has become a common post-training algorithm of large language models~\citep{lu2025onpolicydistillation}. In OPD, the student samples its own trajectories and the teacher scores every token, providing a dense, per-token learning signal that avoids both the exposure bias of supervised fine-tuning and the sparse rewards of reinforcement learning~\citep{agarwal2024policy}. Frontier models rely on OPD to consolidate domain-specialized teachers into a single deployable agent~\citep{xiao2026mimo,xu2026deepseek,zeng2026glm,team2026kimi}.

However, because the teacher scores the student's tokens conditioned on the student's own reasoning, OPD is sensitive to the reasoning mismatch between the two models~\citep{li2026rethinking}.
The sampled trajectories are on-policy for the student but ``off-policy'' for the teacher~\citep{huang2026off}, which makes the teacher model drift from its own training distribution after seeing student's partial solutions~\citep{ma2026should}.
This can fundamentally shape the teacher's training signal because even a strong teacher that scores higher on benchmarks can fail to improve the student after OPD~\citep{li2026rethinking}.
So far, however, this mismatch has been studied almost exclusively through the lens of task accuracy, leaving the discussion of its effects on token efficiency untouched.

\begin{figure*}[t]
    \centering
    \includegraphics[width=0.88\textwidth]{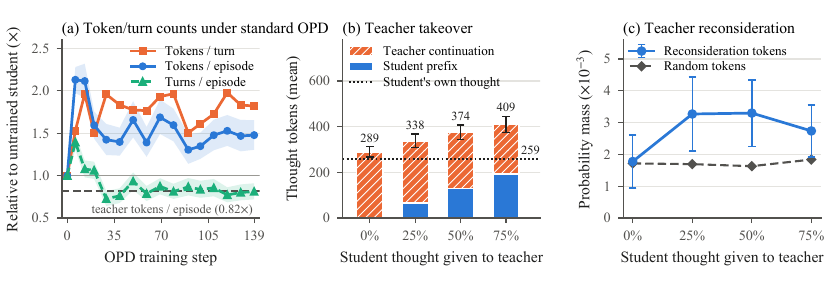}
    \caption{\textbf{OPD lengthens the student's reasoning, and reading the student's thought changes how the teacher continues it.}
    (a) A Qwen3.5-4B student trained with standard OPD from a Qwen3.5-397B-A17B teacher on OfficeQA Pro. Output tokens per episode, turns per episode, and output tokens per turn are shown relative to the student's base model. The dashed line marks the teacher's own tokens per episode.
    (b) The teacher continues the student's turn after reading 0--75\% of the student's thought. Each bar stacks the mean student prefix and the mean teacher continuation, so its height is the mean total thought. The dotted line marks the mean length of the student's own thought.
    (c) The teacher's next-token probability mass on reconsideration tokens (e.g., ``wait'', ``hmm'', ``actually'') over its 24-token continuation, compared with a random token set of the same size and the same probability mass at the start.}
    \label{fig:intro}
\end{figure*}

Our study focuses on multi-turn knowledge-work agents~\citep{opsahl2026officeqa,patwardhan2026gdpval,ma2024spreadsheetbench}, where token efficiency is critical because end-to-end latency directly governs user experience~\citep{langchain2026state}. We begin with a counterintuitive observation: standard OPD causes the student to generate substantially more tokens per episode than before training, even when the teacher itself is more concise (Figure~\ref{fig:intro}a). This growth comes from longer per-turn reasoning rather than more turns, and stems from how the teacher reacts to reading the student's partial thoughts: instead of concluding them, the teacher tends to hesitate and extend the reasoning (Figure~\ref{fig:intro}b--c), passing that habit on to the student.

To break this loop, we propose \method (\methodshort), which conceals the student's internal reasoning from the teacher so that the teacher judges only what the student visibly outputs and executes, while the student remains free to reason in its own context. By decoupling what the teacher sees from how the student thinks, \methodshort removes the teacher's tendency to prolong unfamiliar reasoning and focuses supervision on the actions that govern task success. Without adding extra data, rewards, or hyperparameters, \methodshort makes the trained student both more accurate and substantially more concise, cutting output tokens per episode by more than half across knowledge-work benchmarks.

Overall, our contributions are:
\begin{itemize}[leftmargin=*]
    \item We identify teacher--student reasoning mismatch as a driver of token inflation in multi-turn OPD, showing that conditioning on student thoughts causes the teacher to extend reasoning rather than conclude it.
    \item We diagnose how standard OPD's training signal reinforces excessive thinking while weakening supervision on the student's visible actions.
    \item We propose \method (\methodshort), a minimal, hyperparameter-free modification to OPD, and demonstrate substantial gains in both token efficiency and accuracy, including across model scales and harder held-out tasks.
\end{itemize}

\section{Muted Distillation}
\label{sec:method}

\begin{figure*}[t]
    \centering
    \includegraphics[width=\textwidth]{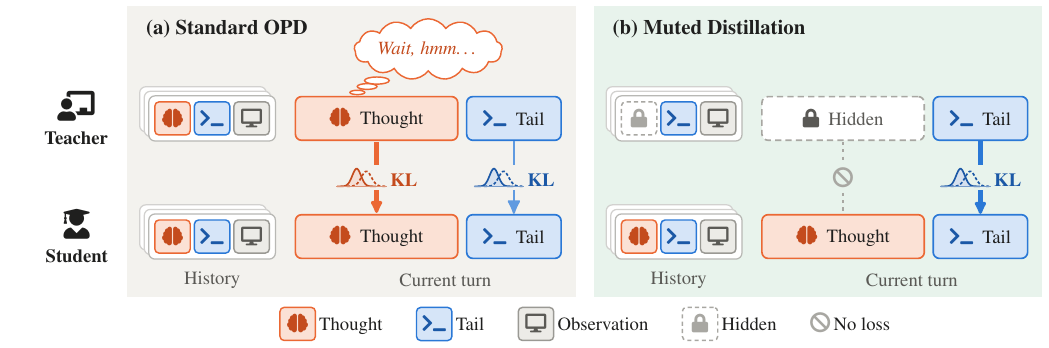}
    \caption{One training turn under each recipe. The teacher (top) scores the student's response (bottom). Arrows mark where the KL signal reaches the student, and their width indicates its strength. The KL icon depicts the teacher (solid) and student (dashed) distributions.
    (a) Standard OPD lets the teacher read the student's thoughts, both in the history and in the current turn, and applies the loss to the thought and the tail. Having read the thought, the teacher tends to extend it rather than conclude it.
    (b) \method hides every thought from the teacher, replacing it with an empty thinking block, and applies the loss to the tail only. The student still generates, and is trained, with its own thoughts in context.}
    \label{fig:method}
\end{figure*}

\subsection{Background: On-policy Distillation}
\label{sec:background}

In on-policy distillation (OPD), a student $\pi_S$ samples a response $y$ for a task input $x$, and a teacher $\pi_T$ scores every sampled token. The student is trained to minimize the reverse Kullback--Leibler (KL) divergence to the teacher on its own samples,
\begin{equation}
\begin{split}
\mathcal{L} = \mathbb{E}_{y\sim \pi_{S}} \Big[ \frac{1}{|\mathcal{T}|}\sum_{t\in \mathcal{T}} & D_{\mathrm{KL}} \big(\pi_S (\cdot\mid x, y_{<t}) \\
&\quad \big\| \pi_T(\cdot\mid x, y_{<t}) \big)\Big],
\end{split}
\label{eq:objective}
\end{equation}
where $\mathcal{T}$ is the set of token positions in $y$. Its gradient takes the form of a policy gradient,
\begin{equation}
    \nabla_S \mathcal{L} = - \mathbb{E}_{y\sim \pi_S} \Big[ \frac{1}{|\mathcal{T}|}\sum_{t\in \mathcal{T}} A_t \nabla_S \log(\pi_S(y_t\mid x, y_{<t}))\Big],
    \label{eq:derivative}
\end{equation}
with the per-token advantage $A_t=\log\pi_T(y_t\mid x, y_{<t}) - \log\pi_S(y_t\mid x, y_{<t})$, which measures how much more likely the teacher is to produce the student's token under the same prefix. Reverse KL is mode-seeking, as the student is penalized for placing mass where the teacher does not, but not for dropping modes the teacher covers~\citep{agarwal2024policy}. Following \citet{lu2025onpolicydistillation}, we estimate the KL from the sampled token only, since the full-vocabulary difference does not fit in memory during the backward pass.

In a multi-turn agentic task, $x$ is the interaction history so far and $y$ is the student's response in the current turn. We write each response as $y = (z, a)$, where $z$ is the thought enclosed in the thinking block and $a$ is the tail, i.e., the visible tokens after \eot, which contain any prose and the command. Let $\mathcal{Z}$ and $\mathcal{A}$ denote the positions of the thought and tail tokens, so that $\mathcal{T} = \mathcal{Z} \cup \mathcal{A}$. Figure~\ref{fig:method}a shows this split under standard OPD, where the teacher reads the whole response and the loss covers both the thought and the tail.

\subsection{Diagnosis: What the Teacher Does with the Student's Thought}
\label{sec:diagnosis}

Standard OPD inflates the student's per-turn thinking even when the teacher itself is more token-efficient (Figure~\ref{fig:intro}a). Examining both the teacher's continuations and the training signal of Eq.~\ref{eq:objective} directly reveals two problems: one with supervising the thought, and one with letting the teacher read it.

{\setlength{\intextsep}{8pt plus 2pt minus 2pt}%
\begin{figure}[H]
    \centering
    \includegraphics[width=0.82\linewidth]{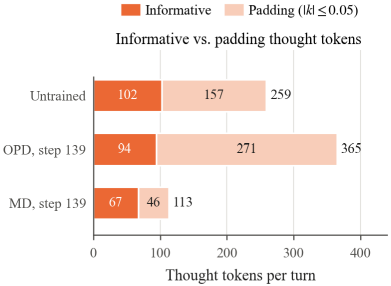}
    \caption{Thought tokens per turn, split into padding (teacher agrees, $|k| \leq 0.05$) and informative tokens, scored with the teacher reading the full turn (OfficeQA Pro training prompts).}
    \label{fig:padding}
\end{figure}

\noindent\textbf{Supervising the thought tolerates padding.}
Figure~\ref{fig:padding} splits the student's thought tokens by the per-token log-ratio $k_t = \log\pi_S(y_t\mid x, y_{<t}) - \log\pi_T(y_t\mid x, y_{<t}) = -A_t$. We call tokens with $|k_t| \leq 0.05$ padding. The teacher already agrees with the student on them, so they carry almost no gradient. After OPD training, padding grows from 157 to 271 tokens per turn while informative tokens stay near 100, so the extra reasoning consists almost entirely of tokens that nothing in the objective pushes back on.
The thought also dominates each update. It holds 79\% of the response tokens, and its gradient norm is a median 5.5$\times$ that of the tail, in a direction nearly orthogonal to it (Appendix~\ref{app:gradients}). Most of every OPD step is therefore spent on a padding-tolerant objective that is unrelated to the student's actions.

\noindent\textbf{Reading the thought encourages extension and weakens action correction.}
Once the teacher reads part of the student's thought, it tends to extend that thought rather than conclude it. Specifically, when we provide the teacher with a 0--75\% prefix of the student's thought and let it continue, the total thought length grows monotonically with the prefix fraction (Figure~\ref{fig:intro}b), and the teacher roughly doubles its probability mass on reconsideration tokens such as ``wait'', ``hmm'', and ``actually'' over its continuation while random baseline tokens are unaffected (Figure~\ref{fig:intro}c). This indicates that the teacher becomes uncertain when reading the student's partial reasoning and tries to extend it, and the mode-seeking reverse KL sharpens the student onto this preference.
The thought also changes how the teacher scores the tail. Figure~\ref{fig:signal} scores the same tail tokens under two teacher contexts. Over the first eight tail tokens after \eot, the per-token KL averages about 0.10 when the teacher reads the thought, against 0.35 when the thought is hidden. Having read the reasoning behind an action, the teacher accepts the action more readily and corrects it less.

\begin{figure}[H]
    \centering
    \includegraphics[width=0.82\linewidth]{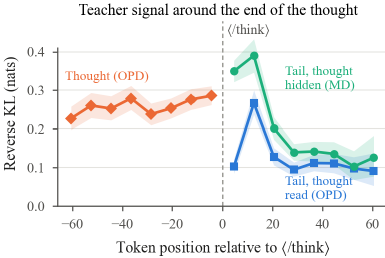}
    \caption{Per-token reverse KL between untrained Qwen3.5-4B student and Qwen3.5-397B-A17B teacher around thought end. Tail curves score the same tokens with thought read (OPD) or hidden (\method) on OfficeQA Pro training prompts.}
    \label{fig:signal}
\end{figure}%
}

Together, the two problems call for two changes: stop supervising the thought, and stop showing it to the teacher.

\subsection{Muted Distillation}
\label{sec:md}

\method makes one change for each problem (Figure~\ref{fig:method}b). It applies the loss to the tail only, and it replaces the student's thought with an empty thinking block $\varnothing$ in the teacher's context. Formally, \method optimizes
\begin{equation}
\begin{split}
\mathcal{L}_{\text{MD}} = \mathbb{E}_{y\sim \pi_{S}} \Big[ \frac{1}{|\mathcal{A}|}\sum_{t\in \mathcal{A}} & D_{\mathrm{KL}} \big(\pi_S (\cdot\mid x, y_{<t}) \\
&\quad \big\| \pi_T(\cdot\mid x, \varnothing, a_{<t}) \big)\Big],
\end{split}
\label{eq:md_objective}
\end{equation}
where $a_{<t}$ denotes the tail tokens preceding position $t$. The teacher conditions on the interaction history $x$, the empty block, and the visible tail. The student's own context is unchanged, so it still generates and is trained conditioned on its full thought. The per-token advantage becomes
\begin{equation}
A_t =
\begin{cases}
\!\begin{aligned}
&\log\pi_T(y_t\mid x, \varnothing, a_{<t}) \\
&\quad - \log\pi_S(y_t\mid x, y_{<t})
\end{aligned} & t \in \mathcal{A}, \\[8pt]
0 & t \in \mathcal{Z},
\end{cases}
\label{eq:md_advantage}
\end{equation}
so the teacher judges each tail token by what the student has done so far, not by how it reasoned, and the thought receives no direct gradient.
\method adds no reward, data, or hyperparameters to OPD, and the teacher scores a shorter context.
Table~\ref{tab:ablation} tests the two changes separately, since mask-think-tail restricts the loss to the tail but still lets the teacher read the thought.

\section{Experimental Setup}
\label{sec:setup}

\subsection{Benchmarks}

We focus on knowledge work agents in this paper because of its need of low latency during deployment, using OfficeQA Pro~\citep{opsahl2026officeqa} and SpreadsheetBench~\citep{ma2024spreadsheetbench}.

OfficeQA Pro is an enterprise question-answering benchmark over U.S. Treasury Bulletins, a corpus of roughly $89{,}000$ pages. Each question requires locating evidence in one or more bulletins. The benchmark has two splits, the 133 Pro questions and 113 easy questions. We train on the easy split only and report all results on the held-out Pro split. The agent works in a sandboxed container. Episodes are capped at 120 turns and 8{,}192 output tokens per turn.

SpreadsheetBench contains 912 spreadsheet manipulation instructions collected from Excel help forums, each paired with about three workbook test cases in which the data differ. A solution is correct on a test case if the designated answer cells of the produced workbook match the golden workbook, and an instruction counts as solved under the hard restriction only if all of the test cases pass. We train on 502 instructions that are disjoint from the 400 instructions verified from experts. The agent sees only the first test case's workbook and must write a reusable solution in its sandbox. The program is then replayed on every test case in a fresh container, and the answer cells are compared with the golden workbook. Episodes are capped at 50 turns, and we report pass@1 under the hard restriction.

\subsection{Evaluation}

On both benchmarks, we sample 8 attempts per task at temperature 1.0 and report pass@1, and the mean number of output tokens per episode. Also, we choose mini-swe-agent as our harness for its simplicity and wide effectiveness~\citep{yang2024sweagent}.

\subsection{Baselines}

\noindent\textbf{Standard OPD.} Our main baseline is on-policy distillation with the objective of Eq.~\ref{eq:objective}, in which the teacher reads the student's whole response and the loss covers both the thought and the tail. It uses the same teacher, training prompts, and hyperparameters as \method, so the two differ only in what the teacher sees and which tokens are trained.

\noindent\textbf{Mask-think-tail.} This ablation keeps the teacher context of standard OPD, so the teacher still reads the student's thought, but applies the loss to the tail only, as \method does. Its gap to \method therefore isolates the effect of hiding the thought from the teacher.

\noindent\textbf{Think-only.} This ablation also lets the teacher read the thought, but applies the loss to the thought only and leaves the tail unsupervised. It is the complement of mask-think-tail and tests what supervising the thought alone contributes.

\noindent\textbf{SFT.} This supervised baseline imitates the teacher. The teacher generates expert trajectories online, and the student is fine-tuned on them, so its training data is off-policy for the student.

\noindent\textbf{Dr.\ GRPO.} We train Dr.\ GRPO~\citep{liu2025drgrpo} as a pure reinforcement-learning reference that learns from the task reward instead of a teacher. It centers each reward on its group mean without dividing by the standard deviation, and normalizes the loss by a fixed length equal to the 8{,}192-token budget. The reward is exact match on OfficeQA Pro and the soft-restriction score on SpreadsheetBench, and each group contains 8 episodes.

\subsection{Models and Training}

We choose Qwen3.5-397B-A17B as the teacher model and Qwen3.5-4B, 9B, and 27B as the student models~\citep{qwen35blog}. The teacher model is served with SGLang~\citep{zheng2024sglang} and used only to score the student's sampled tokens. Our training infrastructure is based on Slime~\citep{slime_github}. All training uses 64k-token training context and at most 8{,}192 tokens per turn. Unless stated otherwise, all methods use Adam with a constant learning rate of $10^{-6}$, weight decay 0.1, sampling temperature 1.0, and one optimizer step per rollout batch, with the per-token loss averaged over all supervised tokens in the batch.

For OPD and \method, each training step samples 16 prompts. We train for 139 steps on OfficeQA Pro and 154 steps on SpreadsheetBench. The 9B and 27B students use the same recipe.

\begin{table}[H]
    \centering
    \caption{Final-checkpoint pass@1 (\%) and output tokens per episode (k) of the Qwen3.5-4B student (step 139 on OfficeQA Pro, 154 on SpreadsheetBench). Icons mark the trained tokens, as in Figure~\ref{fig:method}. Bold marks the best distilled student in each column.}
    \label{tab:ablation}
    \vspace{2pt}
    \small
    \setlength{\tabcolsep}{2.4pt}
    \begin{tabular}{@{}lcrrrr@{}}
        \toprule
        & & \multicolumn{2}{c}{OfficeQA Pro} & \multicolumn{2}{c}{SpreadsheetBench} \\
        \cmidrule(lr){3-4} \cmidrule(l){5-6}
        Method & Loss & Pass@1 & Tokens & Pass@1 & Tokens \\
        \midrule
        \multicolumn{6}{@{}l}{\grouplabel{References}} \\
        \hspace{0.5em}Untrained student & & 13.2 & 24.4 & 30.9 & 14.8 \\
        \hspace{0.5em}SFT & & 30.1 & 30.7 & 49.8 & 24.8 \\
        \hspace{0.5em}Dr.\ GRPO (RL) & & 42.4 & 13.3 & 57.7 & 35.3 \\
        \hspace{0.5em}Teacher (397B) & & 50.9 & 19.9 & 77.3 & 8.6 \\
        \midrule
        \multicolumn{6}{@{}l}{\grouplabel{Teacher reads the thought}} \\
        \hspace{0.5em}Standard OPD & \thinkchip\,\tailchip & 35.0 & 36.0 & 55.5 & 19.5 \\
        \hspace{0.5em}Mask-think-tail & \nochip\,\tailchip & 32.0 & 26.8 & 55.5 & 18.3 \\
        \hspace{0.5em}Think-only & \thinkchip\,\nochip & 30.9 & 27.4 & 57.9 & 16.7 \\
        \midrule
        \multicolumn{6}{@{}l}{\grouplabel{Thought hidden from teacher}} \\
        \hspace{0.5em}\method & \nochip\,\tailchip & \textbf{37.7} & \textbf{15.6} & \textbf{60.3} & \textbf{9.1} \\
        \bottomrule
    \end{tabular}
\end{table}

\section{Results}
\label{sec:results}

\begin{figure*}[t]
    \centering
    \includegraphics[width=\textwidth]{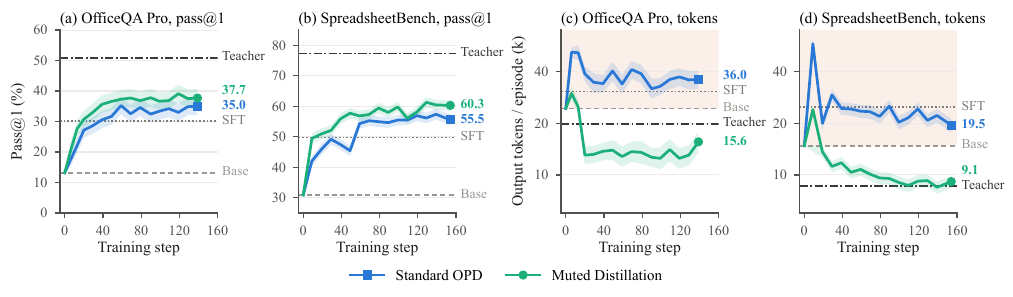}
    \caption{Offline evaluation of Qwen3.5-4B (Qwen3.5-397B-A17B teacher) on OfficeQA Pro and SpreadsheetBench Verified-400.
    Gray lines mark teacher, online SFT, and Base. Final values are labeled, and (c--d) use a log axis shaded above Base.}
    \label{fig:main_curves}
\end{figure*}
\afterpage{%
\begin{table}[t]
    \centering
    \caption{Transfer of the same final checkpoints as Table~\ref{tab:ablation} to the harder OfficeQA Pro v2 and SpreadsheetBench~2.
    Score is pass@1 (\%) on OfficeQA Pro v2 and target-cell accuracy (\%) on SpreadsheetBench~2.
    Bold marks the best trained student in each column.}
    \label{tab:v2}
    \vspace{1pt}
    \footnotesize
    \setlength{\tabcolsep}{2.0pt}
    \begin{tabular}{@{}lrrrrrr@{}}
        \toprule
        & \multicolumn{3}{c}{OfficeQA Pro v2} & \multicolumn{3}{c}{SpreadsheetBench 2} \\
        \cmidrule(lr){2-4} \cmidrule(l){5-7}
        Method & Score & Tokens & Cost & Score & Tokens & Cost \\
        \midrule
        Untrained student & 8.2 & 30.8 & \$11.8 & 26.4 & 25.8 & \$4.1 \\
        \midrule
        Standard OPD & 10.8 & 53.1 & \$18.2 & 34.8 & 29.1 & \textbf{\$4.4} \\
        \method & \textbf{15.1} & \textbf{31.1} & \textbf{\$11.0} & \textbf{37.6} & \textbf{26.0} & \$4.7 \\
        \bottomrule
    \end{tabular}
\end{table}%
}

\noindent\textbf{\methodshort matches OPD's accuracy at far fewer tokens}.
Figure~\ref{fig:main_curves} tracks both students over training and Table~\ref{tab:ablation} reports the final checkpoints.
On accuracy, \methodshort is never behind standard OPD at any evaluated checkpoint. It ends at 37.7 vs.\ 35.0 pass@1 on OfficeQA Pro and 60.3 vs.\ 55.5 on SpreadsheetBench, and it reaches OPD's final accuracy early, within 40 steps on OfficeQA Pro and 60 steps on SpreadsheetBench.
On tokens, the two recipes diverge from the first evaluated checkpoint and never cross. Standard OPD inflates the student, and it peaks at 52k tokens per episode on OfficeQA Pro and stays at 32--41k for the rest of training, 48\% above the untrained student's 24.4k, while \methodshort drops to 13k by step 20 and stays there, ending at 15.6k, 57\% below OPD and 36\% below the untrained student.
SpreadsheetBench shows the same pattern. OPD settles at 20--25k tokens per episode, above the untrained student's 14.8k, whereas \methodshort falls below 10k and ends at 9.1k, 53\% below OPD.
\methodshort is the only distilled student that is more token-efficient than the untrained one, and it approaches the teacher's own budget (19.9k and 8.6k), while standard OPD moves away from it despite distilling from the same teacher.
The SFT baseline trained on the teacher's demonstrations is both less accurate (30.1 and 49.8) and longer than the untrained student (30.7k and 24.8k), so imitating the teacher's traces does not transfer its efficiency either.
Finally, \methodshort also outperforms Dr. GRPO on SpreadsheetBench that is trained with golden verifiers, showing the generalizability of \methodshort.

\noindent\textbf{Gains transfer to harder benchmarks}.
Table~\ref{tab:v2} evaluates the final checkpoints on two harder, held-out variants not used for training or checkpoint selection: OfficeQA Pro v2 (new questions over a separate corpus) and SpreadsheetBench~2 (end-to-end workflows rather than single instructions).
On OfficeQA Pro v2, standard OPD gains only 2.6 points over the untrained student while inflating episodes by 72\% (to 53.1k tokens); \methodshort nearly doubles untrained accuracy (8.2 to 15.1, +4.3 over OPD) at the untrained length (31.1k) and 60\% of OPD's cost.
On SpreadsheetBench~2, \methodshort again leads in target-cell accuracy (37.6 vs.\ 34.8) with 11\% fewer output tokens, though slightly more prompt tokens make cost marginally higher (\$4.7 vs.\ \$4.4).
We report target-cell accuracy because strict whole-task accuracy is under 2\% for all 4B models, where \methodshort still leads OPD (27 vs.\ 16 solved of 2{,}376).
Thus, \methodshort's accuracy gain survives distribution shift, and its token savings hold wherever OPD inflates episodes.

\noindent\textbf{\methodshort's improvements come from not just loss masking}.
Table~\ref{tab:ablation} separates restricting the loss to the tail from hiding the thought.
Mask-think-tail uses the same tail loss as \methodshort while letting the teacher read the thought, so comparing the two isolates muting.
Loss masking alone is a partial remedy: it trims tokens by 26\% on OfficeQA Pro (26.8k) and 6\% on SpreadsheetBench (18.3k), leaving the student 10\% and 24\% longer than before training and losing 3.0 points of pass@1 on OfficeQA Pro.
Hiding the thought removes the remaining inflation and recovers accuracy: \methodshort uses 42\% and 50\% fewer tokens than mask-think-tail with +5.7 and +4.8 pass@1.
Think-only likewise reduces tokens by 24\% and 14\% over OPD, but still inflates reasoning and drops OfficeQA Pro pass@1 to 30.9.
As Section~\ref{sec:method} predicts, a reading teacher sends a weak signal to the tail; muting both strengthens tail supervision and stops the teacher from encouraging extra thought.


\noindent\textbf{\methodshort's advantage holds across student sizes and translates into cost}.
Figure~\ref{fig:pareto} plots each recipe's best checkpoint by accuracy and API cost for 4B, 9B, and 27B students.
At every size, \methodshort matches or beats standard OPD at lower cost, defining the Pareto frontier on both benchmarks: on OfficeQA Pro it cuts cost per 100 episodes by 61\% at 4B (\$3.6 vs.\ \$9.1), 55\% at 9B, and 56\% at 27B while adding 3.9, 1.1, and 3.4 pass@1; on SpreadsheetBench it saves 27\%, 10\%, and 24\% with +3.9, +4.1, and +2.3 points.
First, muting moves a student down a size class: 4B \methodshort matches 9B OPD on OfficeQA Pro (39.1 vs.\ 39.1) at one-tenth the cost (\$3.6 vs.\ \$35.3), and costs less than the untrained 4B model (\$8.8) while nearly tripling its accuracy.
Second, 27B \methodshort overtakes its 397B teacher on both benchmarks (53.9 vs.\ 50.9 and 78.8 vs.\ 77.3) at 31\% and 48\% of teacher cost, whereas 27B OPD falls short (50.5 and 76.5) and costs 1.3--1.6$\times$ more.

\begin{figure}[t]
    \centering
    \includegraphics[width=\columnwidth]{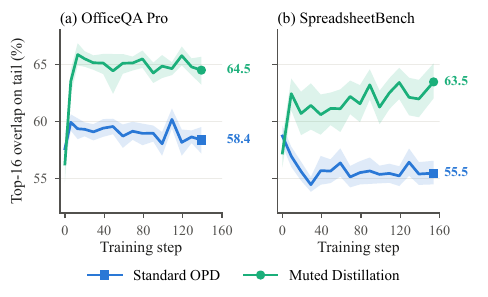}
    \caption{\textbf{\method raises student--teacher agreement.}
    Top-16 candidate overlap between the Qwen3.5-4B student and the Qwen3.5-397B-A17B teacher on student's responses, i.e., the number of next-token candidates shared by both models' top-16 divided by 16, measured at each training checkpoint on that checkpoint's own trajectories.}
    \label{fig:agreement}
\end{figure}

\noindent\textbf{\methodshort raises student-teacher agreement}.
To test whether the student moves closer to the teacher on task actions, Figure~\ref{fig:agreement} tracks top-16 candidate overlap on the student's tail tokens under each recipe's teacher context.
Before training, both contexts agree equally with the student (57.6\% visible vs.\ 56.3\% hidden), giving muting no initial advantage.
Under \methodshort, overlap rises by the first checkpoint and stays high, ending at 64.5\% on OfficeQA Pro (+8.2) and 63.5\% on SpreadsheetBench (+6.3).
Under standard OPD, it barely moves on OfficeQA Pro (58.4\%) and drops on SpreadsheetBench (58.7\% to 55.5\%), so visible actions do not converge to the teacher despite optimizing the loss on every token.
This mirrors Figure~\ref{fig:signal} from the student's side: hiding the thought concentrates teacher disagreement and gradient on the tail, pulling actions toward the teacher, whereas showing it spends disagreement on the thought.
\begin{figure*}[t]
    \centering
    \footnotesize
    \setlength{\fboxsep}{5pt}
    \fbox{\begin{minipage}[t]{0.475\textwidth}
        \textbf{Standard OPD}, turn 6: 776 tokens\\[2pt]
        \textcolor{gray}{\textsc{thought} (723 tokens)}\\
        I see that the table at line 873--893 shows Trust account expenditures data. Looking at the columns, I need to understand what the question is asking for: 1.\ ``investments expenditures'' -- This appears to be the ``Investments in Federal securities \textgreater\ By trust accounts'' column 2.\ ``all other expenditures'' -- This would be everything else in Trust account expenditures \textcolor{gray}{[\ldots\ 642 more tokens of thought omitted \ldots]}\\[3pt]
        \textcolor{gray}{\textsc{prose} (0 tokens)}\\[3pt]
        \textcolor{gray}{\textsc{command} (44 tokens)}\\
        \texttt{grep -Rni "November\textbackslash|December" officeqa\_corpus/\allowbreak treasury\_\allowbreak bulletin\_\allowbreak 1948\_12.txt | grep -i "trust\textbackslash|expenditure" | head -50}
    \end{minipage}}\hfill
    \fbox{\begin{minipage}[t]{0.475\textwidth}
        \textbf{\method}, turn 6: 119 tokens\\[2pt]
        \textcolor{gray}{\textsc{thought} (36 tokens)}\\
        The search didn't find that exact table title. Let me try searching for similar patterns in the 1948 bulletins, particularly looking at different variations of table titles.\\[3pt]
        \textcolor{gray}{\textsc{prose} (43 tokens)}\\
        Let me find the table that contains the monthly data for trust account expenditures. I noticed line 550 has the column headers for investments and other expenditures. Let me look at the full context around that table.\\[3pt]
        \textcolor{gray}{\textsc{command} (30 tokens)}\\
        \texttt{sed -n '540,600p' officeqa\_corpus/\allowbreak treasury\_\allowbreak bulletin\_\allowbreak 1948\_07.txt}
    \end{minipage}}
    \caption{One turn from each final Qwen3.5-4B checkpoint on the same OfficeQA Pro task, at the same stage of the episode. The standard OPD student spends 723 tokens re-reading the question inside its thought and writes no visible explanation; the \method student writes a 36-token thought, states its plan in 43 tokens of visible prose, and issues a more targeted command.}
    \label{fig:qualitative}
\end{figure*}

\noindent\textbf{Thought shrinks after training with \methodshort}.
Figure~\ref{fig:qualitative} shows a turn from each final checkpoint on the same OfficeQA Pro task at the same stage of the episode. The difference is not that the \methodshort student stops reasoning. It still opens a thought, but the thought is short and moves directly to the next action, and the explanation that OPD buries in the thought appears instead as visible prose before the command, where the teacher could see and correct it. The OPD student restates the question at length inside its thought, which the muted teacher never endorsed, and then issues a broad search. This pattern holds for more data. Over all turns with a closed thought at the final OfficeQA Pro checkpoints, the thought shrinks from 540 to 98 tokens on average while visible prose grows from 6 to 267 tokens and commands lengthen slightly from 83 to 130, so \methodshort relocates explanation rather than removing it.

\section{Related Works}

\noindent \textbf{On-policy Distillation.}
On-policy distillation trains the student on its own samples with a per-token signal from the teacher, which avoids the exposure bias of imitating teacher-written traces and the sparsity of outcome rewards~\citep{agarwal2024policy, gu2024minillm,lu2025onpolicydistillation}. It has become a standard stage in frontier post-training pipelines, where it consolidates specialized teachers into one deployable model~\citep{xiao2026mimo,xu2026deepseek,zeng2026glm,team2026kimi}.
A growing line of work studies what happens when the teacher is asked to score trajectories it did not write. The teacher's conditionals drift once it is conditioned on a student prefix~\citep{li2026rethinking,huang2026off}, its ability to teach decays along the sequence~\citep{liu2026prefix}, its stopping behavior can disagree with the student's and inflate responses~\citep{yang2026eos}, and a stronger teacher does not reliably produce a better student~\citep{li2026rethinking,zhao2026capable}.
Proposed remedies mostly act on the teacher or on the target. Training the teacher on student-induced states before distillation~\citep{ma2026should}, updating it jointly with the student~\citep{huang2026off}, filtering or reweighting the per-token signal~\citep{shen2026purified, wang2026not, yang2026prune}, or warming the student up so that its prefixes are closer to the teacher's~\citep{liu2026simple}. Several of these works also report that OPD lengthens responses~\citep{yang2026eos, luo2026demystifying}, but they treat it as an instability to be stabilized rather than as a cost in deployment.
\method is complementary to all of them. It changes neither the teacher nor the loss weighting, only what the teacher is allowed to see, and it needs no additional training or data. The closest relatives are methods that drop reasoning from the distillation target so that the student learns to answer with less or no visible thinking~\citep{kalwar2026masked, kaur2026rethinking}; \method instead keeps the student's thought intact and in its own context, and removes it only from the teacher's.

\noindent \textbf{Token-efficiency during Post-Training.}
Post-training with reinforcement learning tends to lengthen reasoning, and in agentic settings long reasoning can crowd out interaction with the environment~\citep{cuadron2025danger}. Part of the growth is an artifact of the objective. Length-normalized policy gradients reward longer incorrect responses, which Dr.\ GRPO removes with a constant normalizer~\citep{liu2025drgrpo}.
Most approaches to efficiency then add an explicit length objective: budget- or length-conditioned training~\citep{aggarwal2025l1, muennighoff2025s1}, length penalties and long-to-short distillation~\citep{team2025kimi, luo2026o1, hou2025thinkprune}, or adaptive policies that decide when to think at all~\citep{zhang2025adaptthink}. These methods trade accuracy for length along a curve chosen by a hyperparameter, and they presuppose that the training signal itself is length-neutral.
Our setting is different in that the extra tokens come from the distillation signal, which encourages the student's continued reasoning once the teacher has read it. \method removes that source rather than penalizing its symptom, and it does so without a length term, so the student ends up shorter than before training while gaining accuracy. An explicit length objective could still be layered on top of it, which we leave to future work.

\section{Conclusion and Discussion}

On-policy distillation suffers from the reasoning discrepancies between teacher and student, while the empirical results on the issue mostly focuses on the accuracy.
We have argued in this paper that such a mismatch can also cause the token costs to increase significantly on multi-turn knowledge work agents.
Specifically, a Qwen3.5-4B student distilled from a teacher that is more token-efficient than itself ends up thinking roughly twice as long per turn. Our teacher-continuation experiment reveals that the teacher, once it reads the student's partial thought, prefers to extend that thought rather than conclude it. Then reverse KL's mode-seeking behavior encourages this preference. The extra tokens are mostly ones the teacher already agrees with, so nothing in the objective pushes back on them.

\method addresses the issue by replacing the student's thought with an empty block in the teacher's context and gather loss only from the tail tokens, it removes the teacher's opportunity to extend the student's reasoning while leaving the student free to reason in its own context. The change is a few lines in the rollout code and adds no training, data, or hyperparameters. Across OfficeQA Pro and SpreadsheetBench, it matches or improves the accuracy of standard OPD while cutting output tokens per episode by more than half. It also moves 4B, 9B, and 27B students onto the accuracy--cost frontier and we prove that this performance transfers to harder held-out variants of both benchmarks, and brings the student's visible actions measurably closer to the teacher's.

In practice, teacher models are always imperfect, and there are some biases in their produced scores that are irrelevant with the desired capability.
And following Goodhart's law, overoptimizing on such a proxy reward can result in drifting from the true objectives~\citep{gao2023scaling}.
And indeed this explains the numerous reports of empirical failures using the on-policy distillation objective~\citep{fu2026revisiting,zhu2026many,kaur2026rethinking}.
Our study sheds light on the potential of debiasing the teacher model through the delibrate pruning of its contexts.
While in this work we have proved that the teacher model can provide better scores without seeing student's reasoning, we want to point out that a more generic approach might be to selectively prune contexts based on the teacher's preference so that the sample is more ``scorable''.
For instance, a naive approach would be to take the teacher's probability into consideration during student's sampling.
In this way, the samples can potentially induce more accurate scores from the teacher.
Due to the limitated resources, we leave this exploration to the future works.

\section*{Impact Statement}

This paper presents work whose goal is to advance the field of Machine Learning. There are many potential societal consequences of our work, none which we feel must be specifically highlighted here.

\bibliography{main}
\bibliographystyle{icml2026}

\newpage
\appendix
\onecolumn
\section{Training Curves for Larger Students}
\label{app:size_curves}

\begin{figure}[t]
    \centering
    \includegraphics[width=\textwidth]{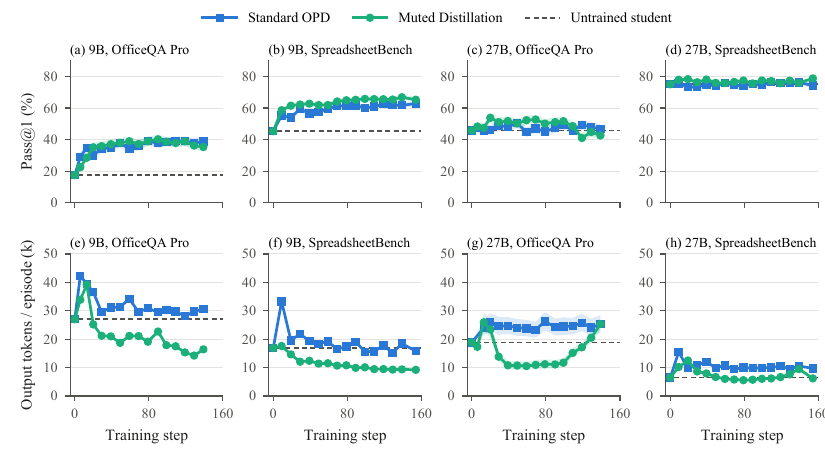}
    \caption{Offline evaluation of every saved checkpoint of the Qwen3.5-9B and Qwen3.5-27B students trained with the Qwen3.5-397B-A17B teacher, under the protocol of Figure~\ref{fig:main_curves}. Shaded bands are 95\% confidence intervals where the evaluation recorded them; the dashed line marks the untrained student.}
    \label{fig:size_curves}
\end{figure}

Figure~\ref{fig:size_curves} shows the per-checkpoint curves behind the 9B and 27B points of Figure~\ref{fig:pareto}. On SpreadsheetBench the picture matches the 4B student: \method is at least as accurate as standard OPD throughout training and ends well below it in tokens (9.2k vs.\ 15.8k at 9B, 6.2k vs.\ 9.7k at 27B), with higher final pass@1 (65.3 vs.\ 62.8 and 78.8 vs.\ 74.3). On OfficeQA Pro the token saving is again large at 9B (16.4k vs.\ 30.7k at the final checkpoint) but the final accuracy is lower (35.3 vs.\ 38.9), and at 27B the two recipes converge: the \method student peaks early (53.9 pass@1 at step 20, the point used in Figure~\ref{fig:pareto}) and then declines, ending at 42.7 pass@1 and 25.3k tokens against 46.6 and 25.3k for OPD. The larger students are therefore where \method is least reliable on OfficeQA Pro, and the best-checkpoint comparison in Figure~\ref{fig:pareto} should be read together with these curves.

\section{Best and Final Checkpoints}
\label{app:best_final}

\begin{table}[t]
    \centering
    \caption{Best and final checkpoints of every standard OPD and \method run with the Qwen3.5-397B-A17B teacher. \emph{Best} is the evaluated checkpoint with the highest pass@1 (the points of Figure~\ref{fig:pareto}); \emph{final} is the last checkpoint of the run (the rows of Tables~\ref{tab:ablation} and~\ref{tab:v2} and the endpoints of Figures~\ref{fig:main_curves} and~\ref{fig:size_curves}). Tokens are mean output tokens per episode in thousands. Three sweeps are incomplete (27B OPD on OfficeQA Pro, 12 of 15 checkpoints; 9B OPD on SpreadsheetBench, 14 of 15; 27B \method on SpreadsheetBench, 10 of 15), so their best checkpoint is the best among those evaluated.}
    \label{tab:best_final}
    \vspace{4pt}
    \footnotesize
    \setlength{\tabcolsep}{3pt}
    \begin{tabular}{@{}llrrrrrr@{}}
        \toprule
        & & \multicolumn{3}{c}{Best checkpoint} & \multicolumn{3}{c}{Final checkpoint} \\
        \cmidrule(lr){3-5} \cmidrule(l){6-8}
        Student & Recipe & Step & Pass@1 & Tokens & Step & Pass@1 & Tokens \\
        \midrule
        \multicolumn{8}{@{}l}{\emph{OfficeQA Pro}} \\
        4B & Standard OPD & 59 & 35.2 & 33.9 & 139 & 35.0 & 36.0 \\
        4B & \method & 119 & 39.1 & 12.5 & 139 & 37.7 & 15.6 \\
        9B & Standard OPD & 79 & 39.1 & 30.9 & 139 & 38.9 & 30.7 \\
        9B & \method & 90 & 40.2 & 22.7 & 139 & 35.3 & 16.4 \\
        27B & Standard OPD & 49 & 50.5 & 24.0 & 139 & 46.6 & 25.3 \\
        27B & \method & 20 & 53.9 & 23.4 & 139 & 42.7 & 25.3 \\
        \midrule
        \multicolumn{8}{@{}l}{\emph{SpreadsheetBench}} \\
        4B & Standard OPD & 139 & 57.4 & 22.2 & 154 & 55.5 & 19.5 \\
        4B & \method & 129 & 61.3 & 9.3 & 154 & 60.3 & 9.1 \\
        9B & Standard OPD & 154 & 62.8 & 15.8 & 154 & 62.8 & 15.8 \\
        9B & \method & 139 & 66.9 & 9.4 & 154 & 65.3 & 9.2 \\
        27B & Standard OPD & 109 & 76.5 & 10.1 & 154 & 74.3 & 9.7 \\
        27B & \method & 154 & 78.8 & 6.2 & 154 & 78.8 & 6.2 \\
        \bottomrule
    \end{tabular}
\end{table}

The main text reports final checkpoints in its tables and best checkpoints in the Pareto plot. Table~\ref{tab:best_final} lists both so that the two can be reconciled. For the 4B student and for every SpreadsheetBench run the choice makes little difference: best and final differ by at most 2.7 points of pass@1 and the \method student is ahead of standard OPD under either reading, with the token gap essentially unchanged. The exception is OfficeQA Pro at 9B and 27B, where the \method runs peak early (steps 90 and 20) and decline afterwards, so the best-checkpoint comparison in Figure~\ref{fig:pareto} is more favorable to \method than the final-checkpoint one in Figure~\ref{fig:size_curves}. We therefore state the size-transfer claim of Section~\ref{sec:results} in terms of the Pareto frontier, which uses the best checkpoint of every recipe including standard OPD, and note the final-checkpoint reversal explicitly in Appendix~\ref{app:size_curves}.

\section{Full Ablation}
\label{app:full_ablation}

\begin{table}[t]
    \centering
    \caption{All trained variants of the Qwen3.5-4B student at the final checkpoint (OfficeQA Pro: step 139; SpreadsheetBench: step 154), same protocol as Table~\ref{tab:ablation}. \emph{Teacher sees} is what the teacher conditions on in place of the student's thought (the thought itself, an empty block, or a thought the teacher writes itself); \emph{student keeps} is whether the student's own sampled thought stays in its context; \emph{loss on} is the set of trained tokens. First-$t$ trains the first 130 (OfficeQA Pro) or 280 (SpreadsheetBench) response tokens.}
    \label{tab:full_ablation}
    \vspace{4pt}
    \footnotesize
    \setlength{\tabcolsep}{2pt}
    \begin{tabular}{@{}lcccrrrr@{}}
        \toprule
        & & & & \multicolumn{2}{c}{OfficeQA Pro} & \multicolumn{2}{c}{SpreadsheetBench} \\
        \cmidrule(lr){5-6} \cmidrule(l){7-8}
        Method & Teacher sees & Student keeps & Loss on & Pass@1 & Tokens & Pass@1 & Tokens \\
        \midrule
        Standard OPD & thought & yes & response & 35.0 & 36.0 & 55.5 & 19.5 \\
        Think-only & thought & yes & thought & 30.9 & 27.4 & 57.9 & 16.7 \\
        Mask-think & thought & yes & command & 28.8 & 31.9 & 55.2 & 19.3 \\
        Mask-think-tail & thought & yes & tail & 32.0 & 26.8 & 55.5 & 18.3 \\
        First-$t$ & thought & yes & first $t$ & 32.1 & 36.0 & 56.7 & 20.6 \\
        Teacher-think & own thought & yes & tail & 25.1 & 39.3 & 49.8 & 16.3 \\
        \midrule
        AOD & empty & no & command & 32.4 & 21.0 & 52.1 & 15.7 \\
        AOD on-policy & empty & yes & command & 33.6 & 14.2 & 50.3 & 12.8 \\
        \method & empty & yes & tail & \textbf{37.7} & \textbf{15.6} & \textbf{60.3} & \textbf{9.1} \\
        \bottomrule
    \end{tabular}
\end{table}

Table~\ref{tab:full_ablation} extends Table~\ref{tab:ablation} with the remaining variants we trained. Three observations support the reading in Section~\ref{sec:method}.
First, every variant in which the teacher reads the student's thought leaves the student above its untrained length (24.4k and 14.8k tokens) on at least one benchmark, including first-$t$, which trains only the opening of each turn.
Second, letting the teacher write its own thought before scoring the tail (teacher-think) is the worst variant on both benchmarks: the teacher's own reasoning is no better a context for judging the student's actions than the student's reasoning, and it is far more expensive to obtain.
Third, among the muted variants the loss set still matters. Removing the student's thought from its own context as well (AOD) or training only the command (AOD on-policy) both lose accuracy relative to \method, which keeps the student's thought and trains the whole visible tail, prose included.

\section{Gradient Analysis}
\label{app:gradients}

\begin{figure}[t]
    \centering
    \includegraphics[width=\textwidth]{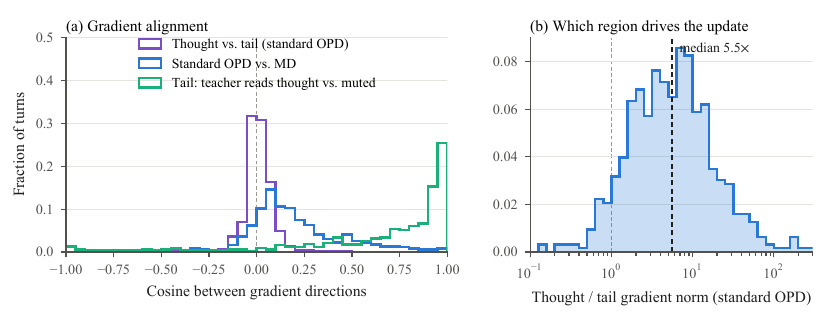}
    \caption{Gradients of the OPD loss on the untrained Qwen3.5-4B student, computed on 630 saved OfficeQA Pro training turns (32 prompts) that contain both a thought and a tail, with the Qwen3.5-397B-A17B teacher. Each turn's gradient is taken over all language-model parameters for the local, unclipped loss of Eq.~\ref{eq:objective} restricted to one token region and one teacher context. (a) Cosine between the thought-region and tail-region gradients of standard OPD; between the full standard-OPD gradient and the \method gradient (Eq.~\ref{eq:md_objective}); and between the tail-region gradient with the teacher reading the thought and with it hidden. (b) Ratio of the thought-region to the tail-region gradient norm under standard OPD.}
    \label{fig:gradients}
\end{figure}

\begin{table}[t]
    \centering
    \caption{The same quantities at the final trained checkpoints (OfficeQA Pro: step 139; SpreadsheetBench: step 154) of the standard OPD and \method students. Each entry is the mean over four fixed four-task batches of saved turns (16 turns per checkpoint), with the range across batches in brackets. The thought is scored with the teacher reading the full turn; ``\methodshort direction'' is the gradient of Eq.~\ref{eq:md_objective} on the same turn.}
    \label{tab:gradients}
    \vspace{4pt}
    \footnotesize
    \begin{tabular}{@{}llcc@{}}
        \toprule
        Benchmark & Checkpoint & Thought vs.\ tail cosine & Standard OPD vs.\ \methodshort direction \\
        \midrule
        OfficeQA Pro & Standard OPD & 0.016 [$-0.010$, 0.038] & 0.157 [0.110, 0.216] \\
        OfficeQA Pro & \method & 0.008 [$-0.012$, 0.018] & 0.115 [0.029, 0.269] \\
        SpreadsheetBench & Standard OPD & 0.011 [$-0.018$, 0.024] & 0.208 [0.023, 0.420] \\
        SpreadsheetBench & \method & $-0.044$ [$-0.094$, 0.002] & 0.121 [$-0.114$, 0.483] \\
        \bottomrule
    \end{tabular}
\end{table}

Section~\ref{sec:method} argues from per-token KL values that the teacher's signal on the thought is weak and that it dilutes the signal on the tail. Gradients let us ask the question directly: how much of a standard OPD update is spent on the thought, and does that part help or hurt the tail? We replay saved turns through the untrained 4B student, recompute the student's log-probabilities with autograd, and take the gradient of the loss of Eq.~\ref{eq:objective} with respect to all language-model parameters, separately for the thought tokens, for the tail tokens with the teacher reading the thought, and for the tail tokens with the thought hidden (the \method gradient of Eq.~\ref{eq:md_objective}). No parameters are updated, so all views share the same student, the same sampled tokens, and the same turns; only the loss mask and the teacher's context differ. Figure~\ref{fig:gradients} summarizes the 630 OfficeQA Pro training turns, and Table~\ref{tab:gradients} repeats the measurement at the final trained checkpoints.

\noindent\textbf{Thought and tail gradients are orthogonal.}
The cosine between the thought-region and tail-region gradients of standard OPD is centered on zero (median 0.007, interquartile range $-0.03$ to $0.05$); 86\% of turns have $|\cos| < 0.1$ and 45\% are negative. Supervising the thought therefore moves the student in a direction that is essentially unrelated to the direction that improves its actions. It neither reinforces nor systematically opposes the tail update; it is a separate objective sharing the same parameters. The same holds at the trained checkpoints of both recipes on both benchmarks (Table~\ref{tab:gradients}, all means within $\pm 0.05$).

\noindent\textbf{But the thought dominates the update.}
The thought holds 79\% of the response tokens in these turns, and its gradient is larger than the tail's in 93\% of turns, by a median factor of 5.5 (Figure~\ref{fig:gradients}b). Consequently the full standard-OPD gradient points mostly along the thought direction: its cosine with the \method gradient is only 0.15 at the median (0.05 to 0.32 interquartile), and 0.12 to 0.21 on average at the trained checkpoints. Under the batch token-mean normalizer of Section~\ref{sec:setup}, standard OPD thus spends most of every step on an orthogonal objective whose per-token signal is the weak, padding-tolerant one of Figure~\ref{fig:padding}. This is the gradient-level form of the dilution argument: the tail signal is not cancelled, it is outweighed.

\noindent\textbf{Muting changes the tail gradient little, and slightly strengthens it.}
The gradient on the tail tokens with the thought hidden from the teacher is well aligned with the gradient on the same tokens with the teacher reading the thought (median cosine 0.83; above 0.5 in 75\% of turns), with a slightly larger norm (median ratio 1.16). Hiding the thought therefore does not replace the tail signal with a different one; it keeps the direction and removes the thought gradient that outweighed it, consistent with the larger per-token KL on tail tokens under the muted view in Figure~\ref{fig:signal}. The turns where the two tail gradients disagree (cosine below zero in 10\% of turns) are the cases in which the teacher's judgment of an action depends on the student's reasoning, which is the behavior \method deliberately forgoes.

These are local, unclipped gradients of single turns at fixed parameters, on 32 correlated prompts, and they do not reconstruct the historical PPO minibatches or Adam-preconditioned updates; the trained-checkpoint pilot uses 16 turns per checkpoint. They describe where the signal goes, not what the optimizer makes of it over training; the training comparisons of Table~\ref{tab:full_ablation} are the test of the consequence.

\section{Where the Token Savings Come From}
\label{app:episodes}

A shorter mean episode could come from failed episodes that give up sooner rather than from more economical solutions. To separate the two, we compare the saved evaluation episodes of the final standard OPD and \method checkpoints (8 attempts per task, temperature 1.0; 4B on both benchmarks and 9B on OfficeQA Pro, 12{,}784 episodes in total) and condition on the outcome. Confidence intervals are 95\% paired task-bootstrap intervals over the evaluation tasks, keeping the eight attempts of a task together; they describe variation across tasks for these trained checkpoints, not variation across training seeds.

\begin{table}[t]
    \centering
    \caption{Token savings conditioned on success, and the length tail, at the final checkpoints of Table~\ref{tab:ablation} and Appendix~\ref{app:size_curves}. \emph{Successful attempts} averages output tokens over successful attempts only, on tasks that both recipes solve at least once, weighting tasks equally; \emph{always solved} restricts further to tasks that both recipes solve in all eight attempts. \emph{At limit} is the share of episodes that end by exhausting the step cap or the context window; P90 is the 90th percentile of output tokens per episode in thousands.}
    \label{tab:episodes}
    \vspace{4pt}
    \footnotesize
    \setlength{\tabcolsep}{2.2pt}
    \begin{tabular}{@{}lrrrcrrrrr@{}}
        \toprule
        & \multicolumn{4}{c}{Successful attempts (tokens, k)} & \multicolumn{1}{c}{Always solved} & \multicolumn{2}{c}{At limit (\%)} & \multicolumn{2}{c}{P90 tokens (k)} \\
        \cmidrule(lr){2-5} \cmidrule(lr){6-6} \cmidrule(lr){7-8} \cmidrule(l){9-10}
        Benchmark, student & Tasks & OPD & \methodshort & Reduction (\%) [CI] & Red.\ (\%) (tasks) & OPD & \methodshort & OPD & \methodshort \\
        \midrule
        OfficeQA Pro, 4B & 73 & 23.7 & 11.3 & 52.3 [45.6, 58.3] & 39.7 (9) & 12.6 & 1.3 & 75.1 & 32.1 \\
        SpreadsheetBench, 4B & 307 & 15.3 & 7.5 & 51.1 [47.5, 54.4] & 36.4 (95) & 7.4 & 3.0 & 53.0 & 20.8 \\
        OfficeQA Pro, 9B & 71 & 22.9 & 11.6 & 49.4 [42.9, 55.0] & 36.0 (12) & 8.7 & 0.9 & 64.3 & 36.3 \\
        \bottomrule
    \end{tabular}
\end{table}

\noindent\textbf{The savings survive conditioning on success.}
Table~\ref{tab:episodes} (left) restricts the comparison to tasks that both recipes solve at least once and to the successful attempts on those tasks. The \method student still uses about half the tokens of the OPD student: 52\%, 51\%, and 49\% fewer on the three comparisons, with confidence intervals that exclude any reduction below 40\%. On the stricter set of tasks that both recipes solve in all eight attempts the saving is 36--40\%, and on SpreadsheetBench these always-solved tasks take essentially the same number of tool steps under both recipes (11.0 vs.\ 11.2) while their tokens fall from 6.3k to 4.0k. The savings are therefore not an artifact of cheaper failures, and on the easiest tasks they come from shorter turns rather than from skipping parts of the tool workflow.

\noindent\textbf{Muting removes the long tail.}
Mean tokens understate the difference at the tail, which is what determines worst-case latency and context pressure. Table~\ref{tab:episodes} (right) shows that the 90th-percentile episode of the OPD student is 2--2.5$\times$ longer than that of the \method student (75.1k vs.\ 32.1k tokens on OfficeQA Pro at 4B), and that the share of episodes ending by exhausting the step cap or the context window falls from 12.6\% to 1.3\% on OfficeQA Pro and from 7.4\% to 3.0\% on SpreadsheetBench at 4B. Much of the OPD tail is unfinished thinking: only 2.1\% of its OfficeQA Pro turns lack a closing \eot, but those turns average about 8.1k tokens each and account for 21.4\% of all output, against 1.2\% of output for \method. In the most extreme saved episode the OPD student emits 244k tokens across 26 turns that each hit the 8,192-token generation limit, revisiting the same table interpretation without producing a usable action, before the context window is exhausted; the \method attempt on the same task is also incorrect but submits after 101k tokens. Such cases are cheaper failures rather than extra successes, which is why the success-conditioned comparison above is the one that supports the efficiency claim.

\section{A Third Benchmark: JobBench}
\label{app:jobbench}

\begin{figure}[t]
    \centering
    \includegraphics[width=\textwidth]{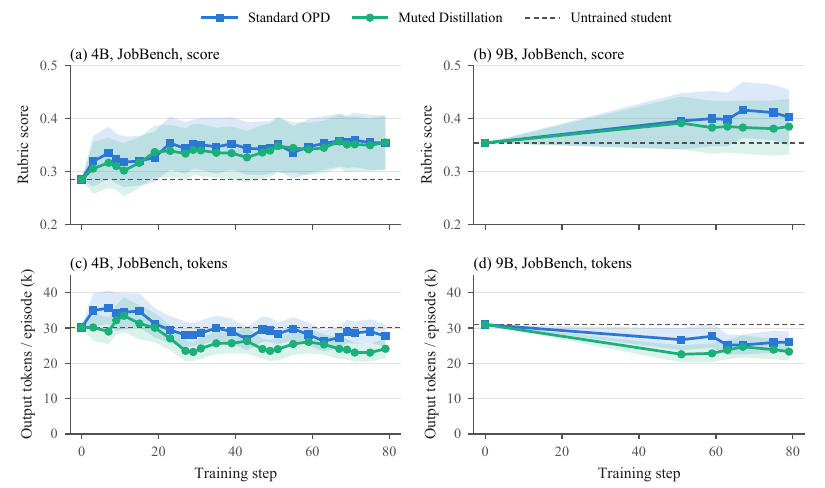}
    \caption{JobBench: offline evaluation of the saved checkpoints of the Qwen3.5-4B and Qwen3.5-9B students trained on the main split with the Qwen3.5-397B-A17B teacher and evaluated on the easy split (8 attempts per task, temperature 1.0, 120-step cap, 95\% task-bootstrap intervals). The 9B run was evaluated from step 51 onward, with an 8k-token per-turn budget; the 4B run uses 16k. The dashed line is the untrained student under the same protocol.}
    \label{fig:jobbench_curves}
\end{figure}

OfficeQA Pro and SpreadsheetBench both have verifiable rewards. To check that the behavior of \method does not depend on that, we also train on JobBench, a benchmark of 128 professional-deliverable tasks in which the agent receives a folder of source documents (spreadsheets, documents, PDFs, and databases; about ten files per task) and a brief from one of 35 white-collar occupations, and must write its deliverables into an output directory. The reward is an LLM judge: each task comes with a weighted rubric, each rubric item passes only if all of its sub-criteria pass, and the score is the normalized weighted sum. We train on the 65-task main split and evaluate on the 63-task easy split, whose scenarios are unrelated and share only the occupation labels, using the same harness as OfficeQA Pro and the same training recipe as Section~\ref{sec:setup}. Grading uses a different judge from the public leaderboard (Gemini~3.5~Flash at low reasoning effort), so absolute scores are not comparable to published numbers; the within-figure comparisons are like for like.

Figure~\ref{fig:jobbench_curves} shows the curves. At 4B the two recipes reach the same score (0.355 for \method vs.\ 0.354 for standard OPD at the final checkpoint, step 79, from 0.285 untrained), while the \method student uses 13\% fewer output tokens (24.1k vs.\ 27.7k) and sits below the OPD student in tokens at every evaluated checkpoint. The saving is smaller than on the two main benchmarks, where standard OPD inflates episodes well above the untrained length; on JobBench both recipes end below it (30.2k untrained), so there is less inflation for muting to remove. For reference, the online SFT baseline at the same step scores 0.342 at 37.7k tokens. At 9B the picture is mixed: \method is again cheaper (23.3k vs.\ 26.0k) but scores lower at the final checkpoint (0.385 vs.\ 0.403, both well above the 0.354 of the untrained student) with overlapping confidence intervals, mirroring the 9B OfficeQA Pro result of Appendix~\ref{app:size_curves}. We read JobBench as evidence that the token saving of muting does not depend on a verifiable reward, and as a second reminder that it does not guarantee an accuracy gain for the larger student.

\section{Combining \method with Reinforcement Learning}
\label{app:rl}

\begin{figure}[t]
    \centering
    \includegraphics[width=\textwidth]{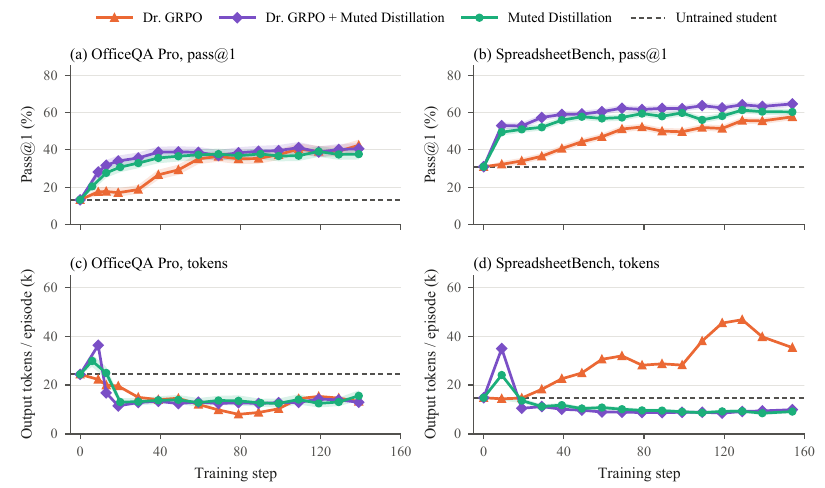}
    \caption{Offline evaluation of every saved checkpoint of the Qwen3.5-4B student trained with the task reward alone (Dr.\ GRPO), with the task reward plus the \method signal, and with \method alone, under the protocol of Figure~\ref{fig:main_curves}.}
    \label{fig:rl_curves}
\end{figure}

\begin{table}[t]
    \centering
    \caption{Final checkpoints of the three recipes in Figure~\ref{fig:rl_curves} (step 139 on OfficeQA Pro, 154 on SpreadsheetBench), and their transfer to OfficeQA Pro v2 and SpreadsheetBench~2 under the protocol of Table~\ref{tab:v2}. Dr.\ GRPO runs use 8 episodes per prompt; \method alone uses one.}
    \label{tab:rl}
    \vspace{4pt}
    \footnotesize
    \setlength{\tabcolsep}{2.4pt}
    \begin{tabular}{@{}lrrrrrrrr@{}}
        \toprule
        & \multicolumn{2}{c}{OfficeQA Pro} & \multicolumn{2}{c}{SpreadsheetBench} & \multicolumn{2}{c}{OfficeQA Pro v2} & \multicolumn{2}{c}{SpreadsheetBench 2} \\
        \cmidrule(lr){2-3} \cmidrule(lr){4-5} \cmidrule(lr){6-7} \cmidrule(l){8-9}
        Method & Pass@1 & Tokens & Pass@1 & Tokens & Pass@1 & Tokens & Score & Tokens \\
        \midrule
        Dr.\ GRPO & \textbf{42.4} & \textbf{13.3} & 57.7 & 35.3 & \textbf{15.6} & \textbf{16.2} & 16.6 & 79.8 \\
        Dr.\ GRPO + \method & 40.6 & 12.9 & \textbf{64.6} & 9.9 & 9.6 & 31.9 & \textbf{38.5} & \textbf{25.1} \\
        \method & 37.7 & 15.6 & 60.3 & \textbf{9.1} & 15.1 & 31.1 & 37.6 & 26.0 \\
        \bottomrule
    \end{tabular}
\end{table}

\method needs no reward, but nothing prevents adding one. We train a variant whose per-token advantage is the sum of the Dr.\ GRPO advantage (the group-centered task reward with a constant length normalizer, Section~\ref{sec:setup}) and the muted distillation signal on the tail, using Dr.\ GRPO's rollout geometry of 8 episodes per prompt. Figure~\ref{fig:rl_curves} and Table~\ref{tab:rl} compare it with each ingredient alone.

The two benchmarks behave differently. On SpreadsheetBench, where the task reward alone drives the student into very long episodes (35.3k tokens at the final checkpoint, 2.4$\times$ the untrained student, and up to 46.8k during training), adding the muted signal removes the inflation and improves accuracy at the same time: 64.6 vs.\ 57.7 pass@1 at 9.9k tokens, which is also 4.3 points above \method alone at nearly the same length. The combination is the best SpreadsheetBench student we trained, and its advantage carries over to SpreadsheetBench~2 (38.5 target-cell accuracy at 25.1k tokens, against 16.6 at 79.8k for Dr.\ GRPO).
On OfficeQA Pro the task reward alone is already short and accurate (42.4 pass@1 at 13.3k), and the combination changes little at the final checkpoint (40.6 at 12.9k), apart from a larger early spike in tokens (36.3k at step 9, against 29.9k for \method alone and none for Dr.\ GRPO); its transfer to OfficeQA Pro v2 is also weaker than either ingredient's. We read this as the two signals pulling in different directions when the reward is informative: the verifiable reward already shapes the tail, and the distillation term adds a second target for the same tokens.

Two caveats apply. These are single runs, and the Dr.\ GRPO recipes use 8$\times$ the rollout compute per step of the distillation runs, so the comparison with \method alone is not compute-matched. The practical conclusion is narrower than the SpreadsheetBench numbers suggest: when a verifiable reward is available, \method can be added to it at no cost in tokens, and it helps most where the reward alone fails to keep episodes short.

\section{Harness and Prompts}
\label{app:harness}

\paragraph{Agent loop.}
Both benchmarks use the same loop. The system prompt (below) and a task message open the conversation. In each turn the model generates a thought and a response; the harness extracts exactly one fenced \texttt{bash} block from the response, executes it in the task's sandbox container, and returns the command's output as the next user message, truncated to 12{,}000 characters by keeping the first and last 6{,}000 and inserting \texttt{[... omitted N characters ...]}. A response without exactly one fenced block receives a fixed format-error message (\emph{Format error: respond with exactly one fenced command block, for example: \ldots}) and costs a turn. Commands time out after 30 seconds. The episode ends when the model calls \texttt{submit}, when it exhausts the turn cap (120 on OfficeQA Pro, 50 on SpreadsheetBench), or when the context budget is exceeded (64k tokens during training; the model's native 262k context at evaluation), with at most 8{,}192 generated tokens per turn.

\paragraph{OfficeQA Pro sandbox.}
A \texttt{python:3.11-slim} container with the parsed corpus mounted read-only at \texttt{officeqa\_corpus/} as one text file per bulletin (\texttt{treasury\_bulletin\_YYYY\_MM.txt}) and a writable workspace containing \texttt{question.txt}. The container has no network access. The task message is:

\begin{lstlisting}
Question: <question text>

Workspace: /workspace
Useful files: question.txt, source_files.txt, manifest.json.
Corpus path: officeqa_corpus/

Search and read the parsed Treasury Bulletin corpus, calculate as needed, then submit the final answer.
\end{lstlisting}

\noindent The system prompt, with the corpus-specific fragments rendered for the Treasury Bulletin corpus, is:

\begin{lstlisting}
You are an autonomous OfficeQA agent and an expert in answering questions related to the U.S. Treasury and economy.

The Treasury Bulletin corpus is available in the workspace at `officeqa_corpus/`. Use filesystem search to locate relevant documents and sections; these documents are long, so do not read the whole corpus or whole files unless needed.

If multiple documents report the same metric, use the most up-to-date revision unless the question specifies an exact date or document. Do not rely on the first matching value you encounter.

You can issue exactly one command per turn. Put the command in exactly one fenced bash block and do not write anything after the block.

Available actions:
- Run shell commands such as `ls`, `find`, `grep`/`rg`, `sed`, `head`, `cat`, and `python3` to search, read evidence, and calculate.
- `submit "<answer>"`: Submit the final answer and finish the task.

Useful corpus patterns:
- Filenames look like `treasury_bulletin_YYYY_MM.txt`.
- Prefer targeted searches such as `grep -Rni -A 3 -B 3 "phrase" officeqa_corpus/treasury_bulletin_1954_*.txt`.
- Use Python for arithmetic, statistics, regressions, and careful unit conversion.

REQUIRED ANSWER FORMAT:
When you have the final answer, call `submit "<answer>"` with only the answer value inside the quotes.

Example:
```bash
grep -Rni -A 2 -B 2 "national defense" officeqa_corpus/treasury_bulletin_1941_01.txt | head -50
```

Example:
```bash
submit "123.45"
```

Use maximum precision in calculations unless the question specifies otherwise. Never ask follow-up questions.
\end{lstlisting}

\paragraph{SpreadsheetBench sandbox.}
A container with Python 3.11, \texttt{openpyxl}, \texttt{pandas}, \texttt{numpy}, \texttt{xlsxwriter}, \texttt{matplotlib}, LibreOffice, and a \texttt{view\_xlsx} helper, mirroring the benchmark's official image, with no network access. Only the first test case's workbook is mounted; golden workbooks and the other test cases never enter the container. The task message gives the instruction, its type (cell- or sheet-level), the answer position, and the fixed paths of the visible input, the visible output, and the reusable program:

\begin{lstlisting}
## SpreadsheetBench v1 task
Instruction: <instruction text>
Instruction type: <Cell-Level Manipulation | Sheet-Level Manipulation>
Answer position: <e.g. D2:D6>

Visible input: /workspace/<input>.xlsx
Visible output: /workspace/<output>.xlsx
Reusable program: /workspace/solution.py

Create a general solution.py, run it on the visible input, verify the output, and submit. The evaluator will run the program on held-out workbook variants.
\end{lstlisting}

\noindent The system prompt is:

\begin{lstlisting}
You are an autonomous spreadsheet automation engineer working in a Linux sandbox.

Your job: solve the spreadsheet instruction by writing a reusable Python program at /workspace/solution.py. The program must read the input path from $SSB_INPUT and save the completed workbook to $SSB_OUTPUT. It will be evaluated on additional workbook variants that have the same task but different values, so never hard-code values copied from this workbook.

You can issue exactly one command per turn. Put the command in exactly one fenced bash block and do not write anything after the block.

Available actions:
- Run shell commands. python3 has openpyxl, pandas, numpy, xlsxwriter, and matplotlib installed; LibreOffice (`soffice`) is available.
- `view_xlsx <file> list`: list workbook sheets.
- `view_xlsx <file> content "<sheet>" [start_row [end_row]]`: inspect values and formulas.
- `submit`: finish. It succeeds only after both /workspace/solution.py and $SSB_OUTPUT exist.

Required workflow:
1. Use the supplied Instruction type and Answer position, then inspect $SSB_INPUT and infer a general solution from labels, formulas, and workbook structure. Answer position identifies the destination cells; it does not contain their correct values.
2. Write /workspace/solution.py. It must use os.environ["SSB_INPUT"] and os.environ["SSB_OUTPUT"], preserve unrelated cells and formatting, and derive results from workbook contents rather than fixed answers.
3. Run `python3 /workspace/solution.py` to produce $SSB_OUTPUT for the visible workbook.
4. Reopen and verify the output, then run `submit`.

Example:
```bash
view_xlsx "$SSB_INPUT" list
```

Example:
```bash
submit
```

Never ask follow-up questions. Golden workbooks and held-out variants are not available in the sandbox.
\end{lstlisting}

\paragraph{Grading.}
On OfficeQA Pro the submitted string is compared with the reference after normalization (Unicode minus signs, whitespace, punctuation, and unit abbreviations); numerical answers match within a 1\% relative tolerance, multi-part and date answers require agreement on both their numbers and their text, and non-numerical answers use fuzzy text matching. On SpreadsheetBench the submitted \texttt{solution.py} is re-run on every test case of the instruction in a fresh container, each output workbook is recalculated with LibreOffice, and the cells in the answer position are compared with the golden workbook; an instruction is solved only if every test case matches (hard restriction). The training reward on SpreadsheetBench is the fraction of test cases that match (soft restriction).

\end{document}